\documentclass[final]{elsarticle}
\usepackage[margin=1in]{geometry}
\usepackage{lineno,hyperref}
\usepackage{amssymb}
\usepackage{graphicx}
\usepackage{amsmath}
\usepackage{makecell}
\usepackage{hyphenat}
\usepackage{numprint}
\usepackage{setspace}
\usepackage{url}
\usepackage{natbib}
\modulolinenumbers[5]

\journal{Expert Systems with Applications}

\biboptions{authoryear}

\begin{document}

\begin{frontmatter}

\title{Marketing Analytics: Methods, Practice, Implementation, and Links to Other Fields}

\author[msuaddress]{Stephen L. France\corref{mycorrespondingauthor}}
\cortext[mycorrespondingauthor]{Corresponding author}
\ead{sfrance@business.msstate.edu}

\author[uwmaddress]{Sanjoy Ghose}
\ead{sanjoy@uwm.edu}

\address[msuaddress]{College of Business, Mississippi State University, \\
114 McCool Hall, 40 Old Main, P.O. Box 5288, MS 39762, USA}

\address[uwmaddress]{Sheldon B. Lubar School of Business, University of Wisconsin-Milwaukee, \\
P.O. Box 742, 3202 N. Maryland Ave., Milwaukee, WI 53201-0742, USA}

\begin{abstract}
Marketing analytics is a diverse field, with both academic researchers and practitioners coming from a range of backgrounds including marketing, expert systems, statistics, and operations research.  This paper provides an integrative review at the boundary of these areas. The aim is to give researchers in the intelligent and expert systems community the opportunity to gain a broad view of the marketing analytics area and provide a starting point for future interdisciplinary collaboration. The topics of visualization, segmentation, and class prediction are featured.  Links between the disciplines are emphasized. For each of these topics, a historical overview is given, starting with initial work in the 1960s and carrying through to the present day. Recent innovations for modern, large, and complex ``big data'' sets are described. Practical implementation advice is given, along with a directory of open source R routines for implementing marketing analytics techniques. 
\end{abstract}


\end{frontmatter}

\section{Introduction}
\label{sec:Introduction}

It is estimated that the worldwide market in business intelligence and analytics will be worth \$200 billion by 2020, up from  \$130 billion in 2016  \citep{REF1149}.  This growth essentially is driven by data \citep{REF1150}.  Large scale corporate databases, mobile-apps, web analytics data, social media, and sensor data, all contribute to what is commonly referred to as ``information explosion''.  Many of the most interesting and practical applications of analytics are in the field of marketing.  In fact, according to an IDG data analytics survey of information systems and analytics executives \citep{REF1151}, the top objective (55\% of respondents) for analytics implementations is to ``improve customer relationships'', a core aspect of marketing.  In addition, the top three challenges for analytics are `` finding correlations across multiple disparate data sources'', ``predicting customer behavior'', and ``predicting product or service sales''.  The second two objectives are direct marketing objectives, while the first objective encompasses a range of consumer analysis applications, including market basket analysis and customer segmentation.  The implementation of techniques to solve these challenges is enabled by the availability of large amounts of marketing data.  For example, the Oracle Cloud includes customer data gathered from a myriad of sources including web browsing behavior, on-line bookings, credit card bookings, scanner purchases, and media viewing habits.  Overall, given the needs, challenges, and available data described above, there is increasing scope for work in marketing analytics, from the perspective of both practitioners and academic researchers.

\section{Objectives}
\label{sec:Objectives}

The purpose of this review is to provide a practical, implementation based overview of marketing analytics methodology. A major objective is to synthesize work from different academic areas.  Contributions in marketing analytics come from a variety of fields including expert systems, marketing science, data mining, statistics, and operations research.  Each discipline has its own literature, and quite often there is little knowledge of similar work in other fields, resulting in some reinventing of the wheel.  Each field has it's own emphasis.  \citet{REF1152} notes that while data mining covers many of the same areas as traditional statistics, there is an emphasis on large datasets, exploring data, pattern analysis, and dealing with poor quality data where the usual statistical assumptions do not hold, such as when there is non-stationary data, sample bias, statistical dependence, and contaminated data.

Analytics research in marketing science has particular niche areas of emphasis, including econometric analysis \citep{REF1153}, Bayesian statistics \citep{REF1154}, and psychometric methods \citep{REF294,REF159}.  Operations research has a particular focus on pricing \citep{REF1155} and location optimization problems \citep{REF1156}.  However, marketing analytics is an area where it is impossible to impose strict disciplinary boundaries.  For example, data mining and statistics have become much closer over the last few years, with more statistical rigor from data mining and machine learning researchers and more emphasis on computational implementations and larger datasets from statisticians.  This is exemplified by statistical learning work \citep{REF257} that develops data analytic techniques in a statistically rigorous manner.  In the commercial arena, the hybrid of data mining and statistics is often referred to as data science.

Another major thread of this review is the increasing importance of expert and intelligent systems in marketing analytics applications.  Expert systems can be thought of as computer systems that embody and improve some aspect of human behavior and decision making processes \citep{REF1580} and utilize both domain knowledge and machine learning or artificial intelligence algorithms \citep{REF1581}.  As early as the 1970s, marketing academics realized that marketing models and methods needed to be implemented as part of integrated decision support systems for business \citep{REF1551}. Subsequently, marketing decision support systems have continued to be an important part of marketing research and have been utilized in a range of marketing areas including promotion planning \citep{REF1582}, services marketing \citep{REF1583}, and product design planning \citep{REF1584}.  The modern expert systems research field is somewhat eclectic and covers a range of application areas and a wide variety of methods from artificial intelligence and machine learning.  However, there is a strong stream of classification research relevant to marketing, and while marketing systems make up the minority of expert systems implementations, a survey \citep{REF1585} found that expert systems in marketing were more effective than those in any other domain.  We hope that with this review, we will provide inter-disciplinary insight and cross-fertilization for marketing analytics researchers working in both expert systems and other arenas.  

A major goal of this review is to provide a technical reference for practitioners and academics wishing to implement marketing analytic techniques.  In each section, a table is given that summarizes some of the software implementations available for the described techniques.  An emphasis is given towards R, as R is open source and is now the de-facto standard platform for statisticians, particularly in the field of data analysis \citep{REF1157}.  More importantly, a global 2015 survey of over 10,000 data science professionals \citep{REF1158} shows that R is the most popular software package, with 76\% of respondents using R, up from 23\% in 2007.  Given the breadth of the marketing analytics field, a review paper cannot cover the whole range of marketing analytics applications in detail.  Thus, a review needs to be selective.  In this review, topics were selected that i) are at the interface of multiple academic disciplines, ii) are commonly utilized in industry, and iii) are scalable to ``big data'' applications.  Given the four Ps of marketing (Product, Promotion, Place, and Price), there is an emphasis towards product and promotion, as these topics are core to academic marketing and marketing analytics.  Some pricing applications are given, but pure operations research revenue and inventory management applications are not.  Likewise, some discussion of ``place'' is given in terms of visualization and social media analytics, but pure location optimization applications are not emphasized.

\begin{figure}[htb]
\centering
\includegraphics[width=16cm]{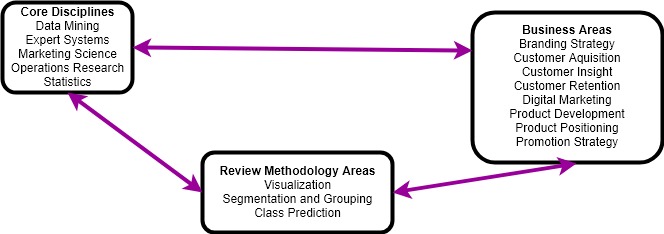}
\caption{Interplay of Basic Discipline, Review Methodology Areas, and Business Areas}
\label{fg:OverallStructure}
\end{figure}

Figure \ref{fg:OverallStructure} gives an overview of the basic disciplines described previously, the methodological areas covered in the review, and the top ten business uses for marketing analytics as defined by the CMO 2016 survey \citep{REF1603} of marketing executives.  Throughout the review, the relationships between these three concepts are emphasized.  Three major topics have been chosen: Visualization, segmentation, and class prediction. These have been chosen because i) they are core to marketing and have been present in the major marketing journals from the 1960s onward, ii) have strong links with expert systems, statistics, data mining, and operations research, and iii) have recently experienced a ``reawakening'' in the academic marketing literature due to the explosion of interest in big data and business analytics.  They are strongly linked, both methodologically and in business applications.  This review does not attempt to give a comprehensive history of analytics in marketing science and some ``purer'' quantitative marketing science topics have been omitted.  An excellent, detailed history of marketing science is given in \citet{REF1573} and a historical discussion on the use of data in marketing is given in \citet{REF1458}.

Other important areas, ommited due to space considerations, such as social network analysis in marketing \citep{REF1459}, recommender systems \citep{REF548}, and time series analysis \citep{REF1460} may be included in a future review.

\section{Visualization}
\label{sec:Visualization}
\subsection{Overview}
Given the previously described problem of ``information explosion'', the ability to understand large, complex, and possibly unstructured data is an important one.  Visualizing data in a parsimonious fashion can be used to help understand patterns in data and to forge new insights.  John Tukey, who helped pioneer the use of data visualization for exploratory data analysis noted that ``The greatest value of a picture is when it forces us to notice what we never expected to see'' \citep{REF685}. To meet increasing needs for visualization, a large number of tools have been developed.  In industry, dedicated visualization packages such as Tableau and PowerBI provide an array of features to allow users to summarize and visualize data, while general providers of business intelligence and analytics software such as SAS, SAP, and IBM have incorporated additional visualization features into their offerings \citep{REF1233}.

\subsection{Foundational Methods}
Visualization research has a long history in marketing. As early as the 1960s, techniques from psychometrics and applied statistics, such as multidimensional scaling (MDS) and factor analysis have been used to visualize marketing data.  Many applications analyzed consumer preferences of products derived from survey data.  For example, given a set products $i=1\dots n$, a proximity matrix ${\bf{\Delta}} = (\delta_{ij})_{\{n\times n \}}$ can be defined, where $\delta_{ij}$ is the proximity between items \textit{i} and \textit{j} and is either elicited directly or derived using a proximity metric from preference data. MDS can use the proximity data to generate a  \textit{p} (usually 2) dimensional output data configuration ${\bf{Y}}= (y_{il})_{\{n\times p\}}$, which can be used to analyze how close products are to one another in the minds of consumers \citep{REF1235}.  If product preferences are available ${\bf{X}} = (x_{il})_{\{n\times m \}}$, where $x_{il}$ is the rating given by user \textit{i} for product \textit{l}, then ``joint space'' mappings \citep{REF1236} can be created that plot both consumers and products on the same map, so that ${\bf{Y}}= (y_{il})_{\{\left(n+m\right)\times p\}}$, where close proximity between consumers indicates similar preferences, and close proximity between a product and a consumer indicates a strong preference for the product from that consumer.  Most multidimensional scaling mapping methods utilize either a decomposition approach or a distance fitting approach.  For classical MDS, ${\bf{\Delta}}$ is converted to a double centered distance matrix ${\bf{B}}={\bf{D}}-{\bf{\bar{D}}}_i-{\bf{{\bar{D}}}'}_i+{\bf{\bar{\bar{D}}}}_i$, where ${\bf{\bar{D}}}_i={\bf{\bar{x}}}_i{\bf{1}}'$ is a matrix of row means and ${\bf{\bar{\bar{D}}}}_i={\bf{\bar{\bar{x}}}}_i{\bf{11}}'$ is a matrix of the overall mean. Then, an eigendecomposition ${\bf{B}}={\bf{Q\Lambda Q'}}$ is performed giving ${\bf{Y}}={\bf{Q\Lambda}}^{1/2}$.  ${\bf{\Lambda}}$ is a diagonal matrix and each entry $\lambda_{ii}$ contains the variance accounted for by the ith new component.  Taking the first $k$ columns of ${\bf{Y}}$  gives the derived \textit{k} lower dimensional solution.

For distance based MDS, ${\bf{\Delta}}$ is transformed to ${\bf{\hat{D}}}$ using a non-metric or metric fitting function, where $\hat{d}_{ij}=F\left(\delta_{ij}\right)$ and ${\bf{Y}}$ is found by optimizing a distance based minimization criterion.  The basic Stress criterion \citep{REF50} is given in Equation \ref{eq:Stress}.  A range of different Stress like criteria and fitting functions are given in \citet{REF277}.
\begin{equation}
\label{eq:Stress}
Stress = \sqrt {\frac{{\sum\limits_i {\sum\limits_{j \ne i} {{{\left( {{d_{ij}}
- {{\hat d}_{ij}}} \right)}^2}} } }}{{\sum\limits_i {\sum\limits_{j \ne i}
{d_{ij}^2} } }}}
\end{equation}
, where $d_{ij}$ is the Euclidean distance between points $i$ and $j$ in the derived solutions.

Many of these early studies note the trade-off between art and science when creating visualizations.  For example, \citet{REF1234} describe the process of rotating \citep{REF1429} the visualization and naming/interpreting the mapping ``dimensions'', either by business interpretation or by some measure of correlation with product attributes \citep{REF1237}.  Extensions to the basic model exist for more complex datasets.  For example, the INDSCAL model \citep{REF157} takes a proximity matrix ${\bf{\Delta}}_s$ for each subject $s=1\dots r$ and creates a single base mapping configuration, but with dimensions weighted (stretched) differently for each individual subject, so that subject \textit{s} on dimension \textit{l} is given weight $w_{sl}$, as shown in \eqref{eq:INDSCAL}.
\begin{equation}
\label{eq:INDSCAL}
\hat{d}_{ij}^s=\sum_{l=1}^{m}\sqrt{w_{sl}\left(x_{il}-x_{jl}\right)^2}
\end{equation}
, where $\hat{d}_{ij}^s$ is the input distance between items \textit{i} and \textit{j} for subject \textit{s}, $x_{il}$ is the output configuration value for item \textit{i} in dimension \textit{l}, and for subject \textit{s} the scaled output values for item \textit{i} in dimension \textit{l} are defined as $y_{il}^{s}=w_{sl}^{\frac{1}{2}}x_{il}$. 

When multiple data sources are present, canonical correlation can be used to combine brand mappings \citep{REF816} into a single composite mapping. MDS techniques give point estimates, which may be unreliable when significant data variability is present. To help visualize the degree of reliability of the points on the MDS maps, several methods have been proposed to estimate confidence intervals for the MDS point estimates. \citet{REF1457} utilizes a maximum likelihood formulation for MDS to create Bayesian credible intervals for points, while \citet{REF1456} estimate mean, variance and correlation data from the weighted INDSCAL $y_{il}^{s}$ values across subjects to create ellipsoidal confidence intervals.

\subsection{Advanced Methods}
There has been a steady stream of product mapping work over the last 30-40 years.  Typically, new methods have been developed to account for new and more complex datasets or more advanced statistical methodology. Multidimensional scaling based product mapping methods have been developed for binary choice data \citep{REF1238}, incomplete preference data \citep{REF1240}, pick any (i.e., choose \textit{k} brands out of \textit{n}) choice data \citep{REF1242,REF1417}, asymmetric brand switching data \citep{REF8}, and scanner data \citep{REF1239,REF154}. If brand data are nominal  then joint space maps can be created using a related technique called correspondence analysis. For example, both yes/no purchase data \citep{REF1250} and nominal brand attribute information \citep{REF1251} can be analyzed in this fashion \citep{REF385}.  

Text data, used for opinion mining and sentiment analysis for brands \citep{REF834}, present a particular challenge, as they are complex and unstructured.  Data can be parsed into a document by feature matrix ${\bf{X}} = (x_{il})_{\{n\times m \}}$, where $x_{il}$ is the count in document \textit{i} of feature \textit{l}.  Here, \textit{m}, the number of so called ``n-gram'' features can be very large and can consist of characters, words, syntactic, and semantic features.  To analyze and visualize the data, feature selection techniques, such as \citet{REF556} can be used to reduce the number of features.  Product attributes can be elicited from the data using text mining methods. \citet{REF1264} extract both attribute information and sentiment polarity information for attributes, then create grid and tree visualizations of product attribute sentiment.  \citet{REF1267} derive product attributes from free-form reviews and then plot these attributes using correspondence analysis. Another approach is Latent Dirichlet Allocation \citep{REF1260}, where each document is considered to consist of a mixture of latent topics and each topic is characterized as a probability distribution over a set of words.  This method is utilized by \citet{REF1261}, who derive topic mixtures for consumer satisfaction from on-line reviews, which they use to create MDS maps, and by \citet{REF1262}, who derive ``service'' and ``user experience'' dimensions for brands from user generated content and plot brands on perceptual maps and radar charts using these dimensions. 

Recent work has applied mapping techniques to newly available internet and social media based data sources, such as online auction bidding data \citep{REF1245}, user generated content \citep{REF1247}, and online reviews \citep{REF1246}. A key component of this work is the creation of proximity values between products from complex and/or unstructured data.  For example, \citet{REF1245} utilize a temporal weighting scheme for bids in which bids by the same user for similar items are given a weight inversely proportional to the time between bids and \citet{REF1246} use text mining techniques to quantify different aspects of writing style that can then be compared using a distance metric. 

Brand mapping techniques are not solely exploratory.  They have particular use in product entry decisions. For example, \citet{REF32} shows how a plot of brand attributes normalized for product cost can be used to determine potential market share for new products and inform brand strategy, an approach that can be applied to scanner data \citep{REF59}.  An alternative approach is to combine a product mapping procedure with the selection of product attributes from conjoint analysis in order to optimally position new products \citep{REF1244}.  In fact, \citet{REF1256}, note that in order for perceptual product maps to be useful as part of an overall product development strategy, there needs to be some relationship between the perceptual maps and the underlying brand attributes.

\subsection{Big Data and Analytics}
While most of the original brand mapping methods were demonstrated on small scale data, the use of web and other large scale data has led to implementations for these data.  For example, \citet{REF1248} introduce a joint-space unfolding decomposition method, which is demonstrated by using ratings from 1000 movie reviewers for 1000 movies to create a joint-space preference map, with reviewers and movie stars plotted on the same map.  \citet{REF1249}, when mapping products using similarities derived from clickstream data, use the approach of creating product maps for sub-markets using an MDS-like criterion function, use a transformation to combine the sub-maps, and then add asymmetry to the data. With the growth in the number of available visualization techniques, methods of evaluating the quality of mappings/visualizations are required.  As most models optimize some measure of goodness of fit to the original data, some ``neutral'' measure of quality is needed.  A common method is to check rank order item neighborhood agreement between the source data and the derived mappings \citep{REF160,REF161,REF406,REF352,REF57}. Solution quality across different neighborhood sizes can be mapped elegantly using the idea of a co-ranking matrix \citep{REF632,REF631}. 

Many innovative techniques have been developed to deal with the specifics of marketing data.  For example, customer relationship management (CRM) \citep{REF655} databases contain large amounts of consumer, contact, finance, and sales information. \citet{REF1253} describe the use of a mapping technique to plot consumers of a general retailer onto a grid using similarities derived from CRM data and to create diagrams shaded by customer attributes related to customer demographics (e.g., age, gender), purchasing behavior (e.g., basket size, spending amount), product category  (e.g., home, outdoors), and product mix.

Beyond mapping, there are many other uses for visualization in marketing analytics. Many of these are extensions to some of the original exploratory data analysis techniques described by Tukey or are adapted from the information visualization literature. For example, in the parallel coordinates approach \citep{REF1255}, high dimensional data are plotted onto a two dimensional map by placing the different data dimensions spaced at equidistant intervals along the x-axis and plotting a line for each item, where the item values are plotted for each dimension and then joined. \citet{REF1257} combine an overall tree structure with a parallel coordinates system in order to visualize a customer survey. \citet{REF1040} use this technique on a longitudinal  scanner dataset to examine the interplay between price and market share for coffee brands by plotting both the market share and price points for the brands over time. In fact, the visualization of data over time can lead to important marketing insights.  For example, \citet{REF39}, examine the price curves for Ebay auctions over time and model commonalities across auctions. \citet{REF1258} visualize and model growth curves for viral video views. \citet{REF1266} describe a pixel-based extension to the bar chart, where for each chart, each brand is plotted in a separate bar and for each customer, the qualitative attributes are coded with different colored pixels.  The use of this technique is demonstrated on sales records for over 400,000 customers.

Most of the applications reviewed thus far give fixed visualizations. As described close to the beginning of the chapter, one of the primary motivations behind visualization is to help users ``explore'', find patterns in, and gain insight from data.  Thus, to be useful to managers, visualization environments must be incorporated into business systems and allow users to explore data interactively.  In fact, in an overview of modern visualization, \citet{REF1272} note that visualization environments for large, complex datasets need to guide users throughout the visualization process and provide tools to sort, select, navigate, annotate, coordinate, and share data. A good example of an interactive system is OpinionSeer \citep{REF1263}, which takes online reviews for hotels and uses opinion mining and subjective logic to create word cloud visualizations paired with perceptual maps that relate opinions to underlying customer categories such as gender, trip type, and age range.  

\subsection{Geographic and Spatial Visualization}
One area of visualization that is particularly pertinent to marketing is geographic visualization and mapping.  Here, the word ``mapping'' generally refers to the overlay of business information onto a cartographic map of physical location.  While the word ``mapping'' in marketing usually refers to perceptual or brand mapping, the concept of applying geographic analysis to marketing problems in not new.  In fact, some early work on multidimensional scaling compared physical maps of location with derived perceptual maps of location, for cities \citep{REF459} and for supermarkets \citep{REF1273}.  In recent years, there has been increasing interest in spatial models in marketing, particularly with regards to econometric models that incorporate spatial or distance effects \citep{REF1275}. For example, \citet{REF1276} build a proportional hazard model on data from an online grocer, which shows that adoption is greatly increased when consumers in neighboring zip codes have also adopted.

Many practical uses of geographic visualization come under the banner of geographic information systems, which are systems that allow users to visualize, explore, and annotate visual data.  A key concept of GIS systems is that of layers.  Data are built up across different layers, with raster layers forming images and vector layers defining features (land boundaries, roads, stores, etc.). GIS systems have been used extensively in retail analytics, having been used to analyze retail location analysis problems, analyze store performance, and plan shopping malls \citep{REF1277}. The retail location problem is an interesting one, as it is at the intersection of several fields.  Original work on gravity models of attraction for stores was given by the Huff gravity model \citep{REF1282}, which has been used extensively in GIS/remote sensing data applications \citep{REF1283}.  The Huff model for the attractiveness of a location is summarized in Equation \eqref{eq:Huff}.
\begin{equation}
\label{eq:Huff}
p_{ij}=\frac{U_jd_{ij}^{\lambda}}{\sum_{k=1}^{n}U_jd_{ik}^{\lambda}}
\end{equation}
Here, $p_{ij}$ is the probability that user \textit{i} uses location \textit{j},  $d_{ij}$ is the distance between user \textit{i} and location \textit{j}, $\lambda$ is a distance decay parameter, and $U_j$ is the utility for location \textit{j}.  In retail location models, $U_j$ is often operationalized as the size of the store.  In the GIS arena and in traditional retail applications, population density data are used to estimate the model.  For facility location, several locations can be analyzed with the respect to the model.  In a retail context, locations can be selected to maximize the market share taken from competitors and to minimize self-cannibalization.  The model is a special case of the Luce choice model \citep{REF1281}. A large number of extensions have been developed for the Huff gravity model, include those that incorporate store image \citep{REF1284}, elasticities of demand \citep{REF1280}, and social media location data \citep{REF1453}.  \citet{REF1285} note that many consumers shop outside of their residence area and describe a model that captures customer flow across geographic areas.  

There has been a tradition of building ``location analysis'' \citep{REF1156} optimization models  for facility location problems in operations research, many which are based on the original Huff gravity model.  Practical location analysis work can combine visual analysis and retail knowledge with operations research methodology.  For example, \citet{REF1278}, in a survey of UK retailers, find that while GIS and quantitative decision support systems are increasingly applied to retail location decision problems, the final choice of location is often manual and local retail knowledge, dubbed ``retail nose'', is important. \citet{REF1279} provide a case study of using GIS for location decisions.  They utilize GIS to analyze customer density and retail competition, using kernel density estimation to identify possible retail sites and combine this analytic work with an overall decision analysis methodology.

\subsection{Software}
\begin{figure}[h]
\centering
\includegraphics[width=16cm]{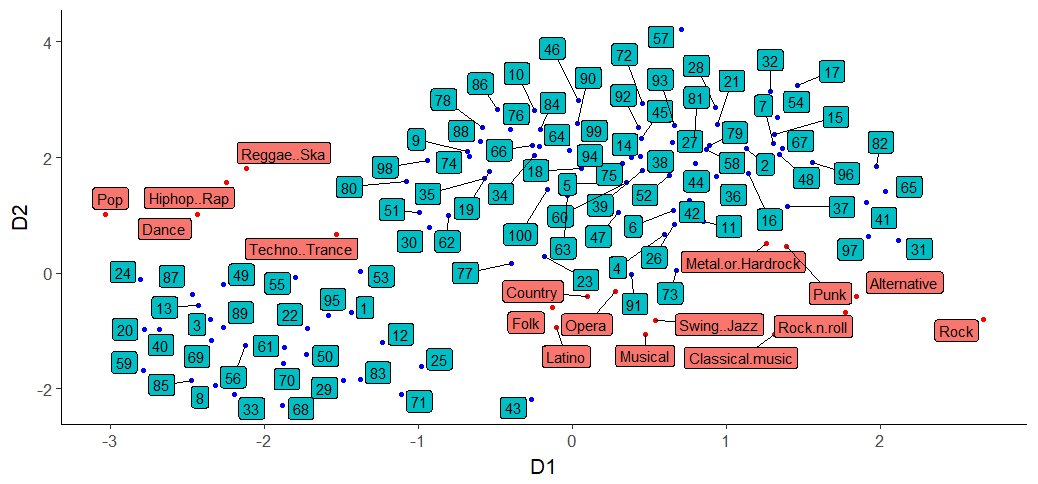}
\caption{Joint Space ``smacof'' Unfolding Solution for Youth Music Preferences}
\label{fg:JointSpacePlot}
\end{figure}

R packages for visualization are listed in Table \ref{tb:RVisualization}.  The packages listed are chosen to present a range of basic methods such as PCA and MDS, along with more complex methods designed for larger datasets, mixed measurement types, and textual data.  In some cases, particularly with methods developed in marketing/OR, software for cited material is not available, so similar methods have been chosen that can best implement the described material. GIS methods are not included, as a large number of interlinked packages are required, but there are several excellent books and tutorials for implementing GIS analyses in R \citep{REF1452,REF1451}.
\begin{table}[h]
\centering
\caption{Visualization Packages in R}
\label{tb:RVisualization}
\begin{tabular}{p{1.9cm} p{3.2cm} p{10.0cm}} \hline
\textbf{CRAN}&\textbf{Reference}&\textbf{Description}\\  \hline
base&None&cmdscale() implements classical multidimensional scaling.\\
base&None&prcomp() and princomp() implement principal components analysis.\\
ca&\citet{REF1435}&Implements simple and multiple correspondence analysis, along with tools for plotting solutions.\\
dimRed&\citet{REF1434}&Interface for dimensionality reduction techniques, including PCA, MDS, ICA (independent component analysis), and techniques for nonlinear data.\\
FactoMineR&\citet{REF1427}&General data mining library containing principal components analysis, correspondence analysis, and multiple correspondence analysis.  Includes measures for mixed measurement types.\\
irlba&\citet{REF1448}&PCA for large, sparse datasets of the type found in word count data and online review data.\\
kohonen&\citet{REF1433}&Methods to implement and visualize SOM (self organizing maps).\\
MASS&\citet{REF1437}&Contains ca() for correspondence analysis, isoMDS() for MDS, and parcoord() for parallel coordinates plot.\\
PCAmixdata&\citet{REF1446}&Methods for mixed data including mixed PCA.\\
smacof&\citet{REF452}&Implements distance MDS, INDSCAL, and joint space (unfolding methods).\\
SpatialPosition&\citet{REF1449}&Implements the Huff model and associated distance based models.\\
syuzhet&\citet{REF1432}&Connects to a range of sentiment analysis parsers, so text can be scored in terms of sentiment and emotional content.\\
tidytext&\citet{REF1431}&Contains framework to parse free text into item $\times$ feature representation needed for most visualization techniques.\\
topicmodels&\citet{REF1430}&Fits topic models using latent dirichlet allocation.\\ 
vegan&\citet{REF1428}&Includes functions for MDS, rotating and interpreting MDS solutions, and cannonical correlation analysis.\\ \hline
\end{tabular}
\end{table}
An example is given in Figure \ref{fg:JointSpacePlot}, where a joint space plot was created using ``smacof'' for a sample of data on youth preferences\footnote{Data can be found at \url{https://www.kaggle.com/miroslavsabo/young-people-survey}} for different movie genres.  One can almost see two distinct clusters of music genres, with the different individuals numbered 1 to 100 positioned relative to the genres. This configuration has some face validity. For example, rock music, punk, alternative, rock `n roll and metal, are all types of guitar driven music performed by bands and are clustered together at the right of the diagram.  Pop, dance, hip-hop, reggae, and techno music tend to be driven by electronic sounds and beats and are clustered together at the left of the diagram.

\subsection{Conclusions}
In summary, much has changed since the initial marketing visualization and mapping research in the 1960s and 1970s.  Computers have become more powerful, datasets have become more complex, and methodology for analyzing data has become more advanced.  However, there are certain research commonalities, which have remained over time.  Visualization will always remain a combination of art and science.  While quantitative techniques can guide interpretation, there is still a need for managerial insight in order to make business decisions from visualizations.  There is a degree of art in how visualizations are pieced together to form a narrative, a process called visualization storytelling \citep{REF1268}. This trade-off between art and science is common to all areas of marketing.  As early as the 1960s, \citet{REF1270} noted that while marketing academics developed scientific theory, the implementation of this theory would require ``art'' based on practitioner experience. Subsequently, the trade-off and tension between art and science in marketing has been a constant topic of interest over the last 30-40 years \citep{REF1271}.  In this respect, visualization is very typical of marketing practice.

\section{Segmentation and Grouping}
\subsection{Overview}
Segmentation is a core activity for marketers. In academia, the idea of segmentation arose in the 1950s. \citet{REF1286} defined market segmentation as the process of ``viewing a heterogeneous market as a number of smaller homogeneous markets in response to differing product preferences'' and notes the interplay between market segmentation and product differentiation strategies.  Initial attempts at segmentation utilized both demographic and psychographic \citep{REF1358} variables.

From an analytics perspective, segmentation refers to the process of grouping or splitting items using a range of segmentation criteria or bases. While the term, ``market segmentation'' implies the segmentation of consumers or products in a market, any meaningful business entity can be segmented, including countries \citep{REF1289}, sales territories \citep{REF1290}, and employees \citep{REF1288}. Segmentation is now an established part of the marketing management literature.  \citet{REF1287} summarize work over prior decades and notes that segmentation can be carried out on a wide range of demographic, psychographic, geographic, and behavioral variables and that in order to be managerially useful, segments need to be substantial, measurable, accessible, and actionable, and differentiable.

\subsection{Foundational Methods}
In many ways, the development of segmentation methods is closely linked to the development of the visualization methods described in the previous section, with authors publishing in both areas. \citet{REF1292} give an application of clustering market territories and describe how the method of cluster analysis can be used to create market partitions.  As with MDS, cluster analysis requires input distances or dissimilarities between items. \citet{REF1291} describe a range of metrics including the Euclidean metric for continuous data and the Tanimoto or Jaccard metric for categorical data, provide additional examples of segmentation for audience profile analysis, customer brand loyalty, experimental gaming, and inter-brand competition, and show how clusters can be overlaid onto MDS solutions. 

Two major clustering methods were explored in this early work.  The first is hierarchical clustering \citep{REF1295}, where a tree structure is derived by either starting at a single cluster and repeatedly splitting the cluster(s) until every item is in its own cluster (divisive clustering) or by doing the opposite (agglomerative clustering) and taking each individual item and combining until there is a single cluster.  For agglomerative clustering, given a distance matrix between items of ${\bf{D}} = (d_{ij})_{\{n\times n \}}$, at each stage of the algorithm the two clusters with the lowest value of $d_{ij}$ are combined.  If two items \textit{a} and \textit{b} are combined then the distance between the newly combined item and a third item \textit{c} can be $min\{d_{ac},d_{bc}\}$ (single linkage), $max\{d_{ac},d_{bc}\}$ (complete linkage), or $\frac{d_{ac}+d_{bc}}{2}$ (average linkage).  There are a multitude of additional schemes, including Ward's method \citep{REF1296} to minimize variance, and centroid clustering, where cluster means are explicitly calculated.  Each combination scheme has its positives and negatives. For example, single linkage clustering produces long drawn out clusters and is susceptible to outliers \citep{REF1297} and complete-linkage clustering produces more spherical clusters.  Various Monte Carlo studies have been run to compare the performance of hierarchical cluster analysis algorithms.  It is generally believed that Ward's method gives the most consistent performance \citep{REF1587}, but some studies have shown that average linkage and centroid clustering algorithms can give better cluster recovery, particularly in cases where there are outliers \citep{REF1589}.

A second type of clustering is partitioning clustering.  Here the items are directly clustered into some predefined \textit{k} number of clusters. Consider a data matrix ${\bf{X}} = (x_{il})_{\{n\times m \}}$, where each row contains the observations for a single item to be clustered. The most common method of doing this is by minimizing the within cluster sum of squares criterion (SSW) given in \eqref{eq:DistClust}.  
\begin{equation}
\label{eq:DistClust}
SSW=\sum_{i=1}^{k}\sum_{{\bf{x}}_j \in C_i}\|{\bf{x}}_j-\bar{\bf{x}}\|^2
\end{equation}
Here, the total Euclidean distance between each row vector ${\bf{x}}_j$ and its centroid $\bar{\bf{x}}$ (average value of items in its cluster) is minimized.  A common method of minimizing this criterion is through the k-means algorithm \citep{REF1298}, where items are randomly assigned to initial seed clusters, the cluster centroids are calculated, and then at each iteration of the algorithm, each item is assigned to its nearest centroid and the centroids are updated.

The authors of this early work noted several challenges.  These include i) how to decide on the number of clusters, ii) how to operationalize distance or similarity measures, particularly in cases when different dimensions/attributes have different scales, iii) how to handle data where dimensions are correlated, and iv) how to properly define the boundaries of the clusters.  Subsequent papers test the use of a large number of metrics including city-block, variance adjusted (Mahalanobis),  and ordinal metrics \citep{REF1294}. \citet{REF1293} develops a metric that accounts both for variation and for weighting dimensions by managerial intuition.   \citet{REF922} note a separate challenge, that of cluster validation.  It is possible that cluster solutions are unstable and change with minor changes in the dataset.  Thus metrics to ensure the stability of the cluster analysis with respect to data variation are needed.  A range of metrics have been developed to test cluster validity \citep{REF1591}; the most common of which is the Hubert-Arabie adjusted Rand index for comparing partitions \citep{REF169}.

\subsection{Managerial Intuition}
While initial marketing analytics work on cluster analysis concentrated on the development of algorithms, managerial intuition was not ignored.  Cluster analysis can be used to split customers into homogeneous subsets based on their characteristics, but does not provide guidance on how to utilize these characteristics as part of an overall marketing strategy.  Several methods were developed to address these challenges. \citet{REF1300} note that there are two typical approaches for segmentation strategy. The first is ``a-priori'' segmentation, where there is some cluster defining descriptor, such as a favorite brand or brand category. The second is post-hoc segmentation, where analytic segmentation is carried out on a range of demographic, behavioral, or psychographic characteristics.  The results of the segmentation are then analyzed with respect to the original segmentation bases (e.g., mean income, brand awareness, etc.) with an eye towards managerial action.  However, \citet{REF1300} argue that post-hoc segmentation, while widely used, does not necessarily predict response to future products or services.  Thus, they give a factorial design based segmentation scheme, which simultaneously analyzes consumer categories and sets of desirable product feature categories using a multi-way linear model.  \citet{REF1301} describe how the two-stage process of calculating segments and then assigning resources to segments can be inefficient and build a resource based mathematical programming model to counter this issue.

\citet{REF1299} summarizes early research in segmentation and notes that any segmentation study should i) start with a managerial problem definition and go through research design, data collection, data analysis, and interpretation stages.  The authors also note that the choice of segmentation bases will depend on the application.  For example, appropriate segmentation bases for product positioning studies include product usage, preference, and benefits, for pricing decisions include price sensitivity and deal proneness, and for advertising decisions include benefits sought, media usage, and psychographic variables.  \citet{REF1302} describe practical limits to what is possible with segmentation and give situations where segmentation is not appropriate. These situations include when a market is too small to be profitable, when a few users dominate the market and most marketing effort is targeted to these users, and when a single brand dominates the market and is purchased by all segments of the population.  

Lifestyle segmentation \citep{REF1307} was developed as existing forms of segmentation were found to be insufficient. Demographic segmentation is too broad-scope and does not contain enough information about behavior, segmentation on psychological attributes is not reliable, and specific brand usage segmentation can be too narrow. \citet{REF1310} notes that lifestyle information  contained in the AIO (activities, interests, and opinions) framework \citep{REF1308} can be used to supplement demographic segmentation and gives insight into consumer behavior that cannot be accounted for by either broad-scope demographic segmentation or usage segmentation on specific products.  \citet{REF1309} describes limits to lifestyle segmentation in that if a segmentation solution is too abstract then it is managerially useless, but conversely, if it is too specific, then it is too close to actual behavioral data to give any additional insight.  Here, the author gives an example of applying lifestyle segmentation to profiles of heavy users vs. non-users of shotgun ammunition. They note that the biggest differentiator between these categories is an enjoyment of hunting, which is closely tied to actual hunting behavior.

\citet{REF1303} posit that any one type of market segmentation cannot be a silver bullet.  All types of segmentation have drawbacks.  A purely demographic segmentation is limited in predicting marketing response.   Psychographic tests adapted to marketing are designed to test underlying psycho-social traits, and have limited ability to predict purchases of specific products.  Behavioral data, such as brand loyalty, can be used to group consumers using purchase patterns, but cannot distinguish between a consumer who buys a product because it is the only product available and a consumer who has high utility for the product.  Overall, \citet{REF1303} conclude that segmentation is only useful when it covers all aspects of consumer and buyer behavior and if segments respond differently to a firm's marketing efforts. This idea is empirically tested by \citet{REF1304}, who carry out a large scale survey of tourism preferences, finding that different segments had different sensitivity elasticities to different advertising strategies.

\subsection{Beyond Partitioning Clustering}
The partitioning clustering methods described thus far assume that all consumers or brands belong to one and only one cluster.  When it comes to real life interpretation of marketing data, this is not necessarily a realistic assumption. Consider a segmentation of movies based on customer preferences, where segments have been found for ``romance'', ``action'', ``horror'', ``sci-fi'', ``teen'', and ``comedy'' movies. A movie may belong in multiple segments; for example, the popular teen movie Twilight could feasibly belong in the ``romance'', ``sci-fi'', and ``teen'' genres.  From a modeling standpoint, define a \textit{k} cluster solution ${\bf{P}} = (p_{ij})_{\{n\times k \}}$, where $p_{ij}\in \{0,1\}$ is a cluster assignment from item \textit{i} to cluster \textit{j}. If $p_{ij}=1$ and $\sum_{j=1}^{k}p_{ij}=1,\forall i=1\dots n$, then this is partitioning clustering.  An alternative way of conceptualizing clustering is overlapping clustering.  Here, each item can be assigned to more than one cluster and $\sum_{j=1}^{k}p_{ij}\geq1,\forall i=1\dots n$.  Overlapping clustering was introduced as a method for product positioning by \citet{REF502}, noting that in benefit segmentation \citep{REF1595}, a product may belong to multiple segments.  For example, chewing gum may be used both as a candy substitute and for dental health.  The authors implement the ADCLUS (ADditive CLUStering) model \citep{REF30}, in which a similarity matrix ${\bf{S}} = (s_{ij})_{\{n\times n \}}$ is decomposed as ${\bf{S=PWP}}'$, where \textbf{P} is the aforementioned overlapping assignment matrix and ${\bf{W}} = (w_{ij})_{\{k\times k \}}$ is a weighting matrix for the different clusters.  The model is used to analyze different usage scenarios for different breakfast food products.  

A number of extensions to the ADCLUS model have been implemented. The INDCLUS model \citep{REF27} allows a similarity matrix for each user or data source and can be used in instances where each user compares multiple products, for example an on-line review scenario. It can also be used when data are split by qualitative attributes, such as region or demographic group \citep{REF55}. INDCLUS can be thought of as a discrete variant of INDSCAL. In the INDCLUS model, for each user $i=1\dots r$, ${\bf{S}}_i={\bf{PW}}_i{\bf{P}}'$, with each user assigned weights in each cluster. \citet{REF4} generalizes INDCLUS for asymmetric data, overlapping or non-overlapping clustering, and a range of weighting options and demonstrates the utility of the model on both brand switching data and celebrity/brand congruence data.  \citet{REF18}, drawing on psychological justification for the use of hybrid models \citep{REF65}, introduce the CLUSCALE model, which contains both continuous and discrete dimensions and is a hybrid of both INDCLUS and INDSCAL models. This model is demonstrated using a segmentation of the car market. 

Yet another clustering conceptualization is that of fuzzy clustering.  Here, $p_{ij}=1$ as per partitioning clustering, but the values of $p_{ij}$ are membership probabilities with $p_{ij}\in [0,1]$.  Fuzzy clustering is useful when dealing with items that are positioned towards the ``edge'' of clusters. For example, consider a customer segmentation where each segment \textit{S} is targeted with a promotion for a specific brand.  However, the customers at the edge of each cluster may have a lower probability of utilizing the promotion and are thus less profitable.  A marketer could save money by only targeting consumers where $p_{ij}$ is above a certain threshold. The most common fuzzy clustering algorithm is a fuzzy variant of the McQueen k-means algorithm entitled c-means clustering \citep{REF1462}. \citet{REF1356} describe a procedure that fuzzifies existing partitioning clusters, to allow extra insight and context to be gained on consumers who are weaker members of segments. \citet{REF941} describes a fuzzy version of the ADCLUS model that can be used to segment both customers and brands. A related model that is somewhat intermediate to k-means clustering and additive clustering is k-centroids clustering \citep{REF1323}.  Here, a consumer $\times$ features matrix ${\bf{X}} = (x_{il})_{\{n\times m \}}$ is decomposed into a binary cluster indicator matrix ${\bf{P}} = (p_{ij})_{\{n\times k \}}$ and a matrix of cluster centroids ${\bf{W}} = (d_{jl})_{\{k\times m \}}$, so that ${\bf{X=PW}}$.  The advantage of this model is a direct correspondence between items (or customers), segments, and features.  To demonstrate the use of the model, a conjoint analysis \citep{REF1325,REF1326} experiment was performed for 600 users and 9 product features, resulting in a $600\times9$ matrix of attribute utilities, which were then input into the segmentation procedure.  Cross-validation and comparison to exogenous variables was used to validate the solutions.

\subsection{Model Based and Econometric Approaches to Segmentation}
Most of the methods described thus far do not make any parametric or distributional assumptions.  This gives advantages in terms of flexibility, but limits some of the output from the procedures in terms of parameter significance and model selection criteria.  Mixture model based clustering \citep{REF944,REF1317,REF1318} approaches that make distributional assumptions have been increasingly utilized for marketing segmentation purposes, particularly over the last few decades.  In mixture model based clustering, the observed data are considered to come from a ``mixture'' of different distributions. Given an observation vector ${\bf{x}}_i$ for each item to be clustered and \textit{k} clusters, the basic mixture model based clustering formulation is given in \eqref{eq:Mixture}.
\begin{equation}
\label{eq:Mixture}
f\left({\bf{x}}_i|{\bf{\Theta}}\right)=\sum_{k=1}^{K}\tau_{k}f_{k}\left({\bf{x}}_i|\text{\boldmath$\theta$}_k\right)
\end{equation}
Here, $f_{k}\left({\bf{x}}_i|\text{\boldmath$\theta$}_k\right)$ is a probability function with parameters $\text{\boldmath$\theta$}_k$ and $\tau_{k}$ is some prior probability of  membership of cluster \textit{k}. Most commonly $\sum_{k=1}^{K}\tau_{k}f_{k}\left({\bf{x}}_i|\text{\boldmath$\theta$}_k\right)$ is implemented as a Gaussian mixture across \textit{k}, with covariance matrix ${\bf{\Sigma}}_k$. Different model assumptions specify different covariance matrices.  The mixture model returns probabilities, which can be scaled as per the membership parameters in fuzzy clustering.  Let $z_{ik}$ be the assignment of item \textit{i} to cluster \textit{k}. Let $CI_i$ be the chosen cluster index for the \textit{i}\textsuperscript{th} item. To get a partitioning solution, the maximum value of $p\left(CI_i=k|{\bf{x}}_i,{\bf{\Theta}}\right)$ is selected.  For a model based overlapping clustering solution, the cluster assignment $z_{ik}=1$ if $p\left(CI_i=k|{\bf{x}}_i,{\bf{\Theta}}\right)>\lambda$, where $\lambda$ is a threshold parameter \citep{REF270}.  

In marketing, model based clustering often comes under the banner of latent class analysis, which can be used to analyze brand switching data \citep{REF15}. In a marketing segmentation context, \citet{REF1319} note that latent class clustering has several advantages over k-means clustering, which include the fact that membership probabilities are generated, which can be used to estimate error, likelihood based diagnostic statistics such as the AIC and BIC \citep{REF1169,REF1171} can easily be used to estimate the number of clusters, a wide range of measurement types are allowed, and exogenous information can be modeled, allowing for simultaneous clustering and descriptive analysis of the clusters.  However, k-means clustering has a range of methods for choosing the number of clusters \citep{REF564}, including the \citet{REF1322} ratio of between to within cluster sum of squares and the gap statistic \citep{REF686}, in which the gap between the sum of squares criterion for the clustering solution and the average sum of squares criterion for clustering solutions from data generated uniformly from the range of the original data is minimized across \textit{k}.  An experimental evaluation of several of these methods is given by \citet{REF1590}. In fact, on a set of experimental data, \citet{REF1321} found that combining k-means clustering with the Caliński-Harabasz method gave better recovery of the number of clusters than model based clustering using AIC and BIC. In addition, variants of k-means have been developed that can account for categorical data \citep{REF1320} and external information.  To summarize, both k-means based and mixture model based styles of clustering have their adherents and advantages/disadvantages.  The technique to be used will depend on the specific dataset and researcher preference.

In a very separate tradition, market share and latent class based methods have been developed by marketing researchers to help explain segmentation behavior and analyze specific market segmentation scenarios for customers and products. Much of this segmentation work builds on the concepts of structured markets and submarkets, where brands within submarkets compete with one another. \citet{REF10} define the concept of submarkets and develop a set of statistical tools to analyze the existence of submarkets based on categorical features, for example \{diesel, gas\} cars, or \{mild, medium, dark\} roasted coffee. A modern extension to identify and visualize brand submarkets is given by \citet{REF1461}.

\citet{REF6} build an explicit latent class model to model brand loyal segments and brand switching segments from brand switching data.  Consider a brand switching matrix ${\bf{S}} = (s_{ij})_{\{n\times n\}}$ that records cross purchases over two purchase occasions, where $s_{ij}$ is the number of households who purchase brand \textit{i} on the first occasion and brand \textit{j} on the second occasion.  The resulting segmentation model is given in \eqref{eq:GSModel}.
\begin{equation}
\label{eq:GSModel}
s_{ij}=\sum_{h=1}^{n+k}\beta_{h}q_{ih}q_{jh}
\end{equation}
Here, there are \textit{n} brand loyal segments and \textit{k} switching segments, $\beta_{h}$ is the size of segment \textit{h}, and $q_{ih}$ is the probability that a consumer in segment \textit{h} purchases brand \textit{i} across multiple time periods.  For brand loyal segments, $q_{ih}=1$ if \textit{h} is the brand loyal for segment \textit{i} and $q_{ih}=0$ otherwise.  The model can easily be fit using a maximum likelihood fitting procedure. There have been many extensions to the core choice-based model to deal with various business scenarios.  These include analyzing segmentation structure over time \citep{REF9}, incorporating promotion effects \citep{REF711}, incorporating price elasticity variables \citep{REF25}, and using multinomial logit models to analyze price sensitivity across purchase incidence and brand choice segments \citep{REF60}.
  
The models described above have found most practical use in analyzing scanner data from supermarkets.  \citet{REF61} note that the ``micro'' household data used to track latent class models allow for more detailed analysis than previously used ``macro'' store level market share data and propose using the segmentation results from latent class analysis to help estimate macro level brand share and momentum parameters.  In addition, \citet{REF1327} shows using a hierarchical Bayesian model that customized store level pricing and promotion strategies derived from micro-level scanner data can improve gross profit margins from 4\% to 10\%.

\subsection{Clusterwise Regression}
As described previously, to be useful, segmentation solutions need to be actionable and some analysis must be made of the resulting segmentation solutions with respect to consumer behavior. One way of examining this is by using a technique called clusterwise regression, described by \citet{REF666}, who expand initial statistical work \citep{REF1316} to create a maximum likelihood model that simultaneously clusters a set of independent variables in a regression, while fitting optimal regression equations relating the independent variables to a dependent variable.  The formulation, given in \eqref{eq:CWRegression}, is similar to the mixture model clustering formulation \eqref{eq:Mixture}, but the mixture is defined across the continuous variable \textit{y}, with each segment $k=1\dots K$ having a regression equation defined by a set of parameters $\text{\boldmath$\beta$}_k$ and a homoskedastic variance $\sigma_k$.
\begin{equation}
\label{eq:CWRegression}
y_i=\sum_{k=1}^{K}\tau_{k}f_{ik}\left(y_i|{\bf{x}}_i,\text{\boldmath$\beta$}_k,\sigma_{k}\right)
\end{equation}
An example is given with trade show data, where the problem was to examine which factors managers considered important for the success of a trade-show visit. Here the dependent variable was the overall rating for a trade show visit, while the independent variables were ratings for  success in certain sub-areas, such as sales, new clients, new product launches, corporate image, morale, and information gathering.  Two distinct segments were found, with one segment prioritizing sales and the other, a more general marketing segment, having more balanced priorities. This approach has been applied to more general customer segmentation problems.  For example, \citet{REF1315} cluster consumers with compulsive behavior shopping problems into two clusters with different drivers for compulsive behavior. 

\citet{REF1305}  extend clusterwise regression to allow for the previously described fuzzy clustering paradigm. A product benefit segmentation example is given in \citet{REF1311}.  Here, based on an MDS configuration derived from a consumer survey, twelve brands of margarine are clustered using ratings on exclusiveness, vegetable content, multiple purpose flexibility, and packaging, with regression used to compare these data with actual product attributes.  As most products could be used for most purposes, a fuzzy clustering solution is more appropriate in this scenario than a partitioning clustering solution. \citet{REF1313} note that within a segmentation context there are often cases where a single dependent variable is not indicative of behavior; for example, when analyzing car repurchase behavior, consumers with a high degree of customer satisfaction are still liable to switch, so to analyze behavior a second variable measuring propensity to brand switch is required. Thus, they develop a multi-criterion programming approach in which clustering is optimized over multiple dependent variables.  \citet{REF1314} note that while clusterwise regression provides a flexible framework for actionable market segmentation, care must be taken, as it is prone to over-fitting, which can be mitigated using a procedure in which the model results are compared to those gained from fitting a model with randomly generated dependent variables.

\subsection{Modern, Large Scale Segmentation Approaches}
As per visualization, over the last 10-20 years, segmentation methods have been developed to both account for increasing large scale, complex, on-line and corporate data. Many of the basic clustering/segmentation techniques described previously have been adapted to deal with this reality. This has generally been achieved by either by improving the efficiency of the implementation algorithms or by parallelization, i.e., splitting up computation into multiple threads and running simultaneously. For hierarchical clustering, \citet{REF1328} develop a method for agglomerative clustering that approximates the process of finding the nearest neighbor using hashing for choosing items to be merged, reducing the complexity of the algorithm.  In addition, there have been several papers on how to best parallelize agglomerative clustering algorithms \citep{REF1329,REF1330,REF1331}. Similar methods exist for k-means clustering. One method is to reformulate the k-means data structure in a tree problem and then develop distance based criteria to prune the tree \citep{REF687,REF1332}, thus reducing the number of possible distance comparisons. In a similar fashion, \citet{REF1333} utilizes the triangle equality to discard distance comparisons that are not possible. As per agglomerative clustering algorithms, there have been several algorithms for parallelizing computation  \citep{REF1334,REF1335}.  A method for combining parallel k-means clustering with the general MapReduce framework for distributed computing is given by \citet{REF1336}.  Model based clustering approaches have been typically constrained by the performance of the EM algorithm \citep{REF1338} used to maximize the likelihood functions.  However, several approaches have been used to speed up estimation. In a basic sampling approach, a random sample of the data is used to calculate the clusters and then an additional ``E'' (expectation) step is used to classify the remaining items.  This approach can be improved by building multiple models from the initial sample and then running through several steps of the EM algorithm to fit the whole dataset to these models \citep{REF1339} or by looking to create new clusters for observations in the full dataset that are fit badly by the sample clusters \citep{REF1340}.  In addition, as per other clustering approaches, parallel methods have been developed \citep{REF1341,REF1343}.

Recent segmentation research has applied a broad range of algorithms to modern datasets.  Of particular interest are neural network algorithms.  These algorithms have in-built parallelization and can be applied to a range of segmentation and classification scenarios. Neural networks take a set of problem inputs and use a hidden layer to transform the inputs to a set of output variables.  Neural networks are applied to a simple product segmentation problem in \citet{REF1344}. The hidden layer equation \eqref{eq:HiddenLayer}, uses a multinomial logit formulation similar to \eqref{eq:GSModel}.
\begin{equation}
\label{eq:HiddenLayer}
s_{il}=\frac{exp\left(\sum_{j=1}^{m}\alpha_{jl}x_{ij}\right)}{\sum_{h=1}^{k}exp\left(\sum_{j=1}^{m}\alpha_{jh}x_{ij}\right)} 
,\quad\hat{y}_{ij}=\frac{1}{exp\left(-\sum_{h=1}^{k}\beta_{hj}s_{ih}\right)}
\end{equation}
Here $s_{il}$ is the segment probability for item \textit{i} in segment \textit{l}, $\alpha_{jl}$ is the weight for feature \textit{j} in segment \textit{l}, and $x_{ij}$ is the value of feature \textit{j} for item \textit{i}.  The fitted output $\hat{y}_{ij}$ for item \textit{i} in feature \textit{j} is calculated across all values of the weighted segment probabilities $\beta_{hj}s_{ih}$ and the neural network error function minimizes the error between each $\hat{y}_{ij}$ and its segment average. A neural network is applied to real world retail segmentation exercises in \citet{REF1345}. The authors utilize an error function that forces probabilities of cluster membership towards 0 or 1 and as per k-means, minimizes the distances to the cluster centroids \citep{REF1346}. The authors found that this method had less reliance on the starting cluster centroid configuration and lower overall error than both k-means and model based clustering.  Neural networks have been used for a range of segmentation applications including on-line shopping behavior \citep{REF1350}, web-log sequence mining \citep{REF1349}, tourism visitor segmentation \citep{REF1348}, and visit/usage segmentation at a dental clinic \citep{REF1347}.  \citet{REF1359}, in a survey of segmentation with cluster analysis, note that there are now a range of different neural network types that have been applied to cluster analysis, including topology representing networks (TRNs), Self-organizing maps (SOMs), and Hopfield-Kagmar (HK) neural networks. Each method has advantages and disadvantages.

A range of other previously described techniques have been adapted for modern datasets. \citet{REF1352} segment a fashion database of 7000 consumers using k-medoids clustering (a variant of k-means with an $L_1$ city-block distance) and subset mining, which is designed to find interesting relations for items/customers who have an specific distribution of a target variable.  In this case, it was used to elicit fashion preferences for overweight consumers.  \citet{REF1594} segment customer visits rather than customers.  This type of analysis allows products and product classes to be matched to different types of purchase occasions, including breakfast, light meal, extended visits around food, and detergents and hygiene. \citet{REF1596} give a modern analysis of benefit segmentation, testing a range of different distance metrics and clustering methods.  The authors find that SOM and fuzzy clustering give strong clustering solutions and that the Gower distance, where data are range scaled within dimensions and then added up as per the Manhattan distance, and the generalized distance metric, which is a measure of generalized correlation, give good results when segmenting on ordinal Likert scale data.

In the marketing literature, over the last 20 years there has been an influential work at the intersection of marketing and accounting/finance on measuring customer lifetime value (CLV) \citep{REF1353} and optimizing marketing processes to maximize this value \citep{REF1354}.  \citet{REF1355} develop a segmentation methodology, which involves calculating CLV and segmenting on CLV along with some measure of customer loyalty.  The methodology is demonstrated on a database of frequent flier information for a Chinese airline.

\subsection{Software}
\begin{figure}[h]
\centering
\includegraphics[width=16cm]{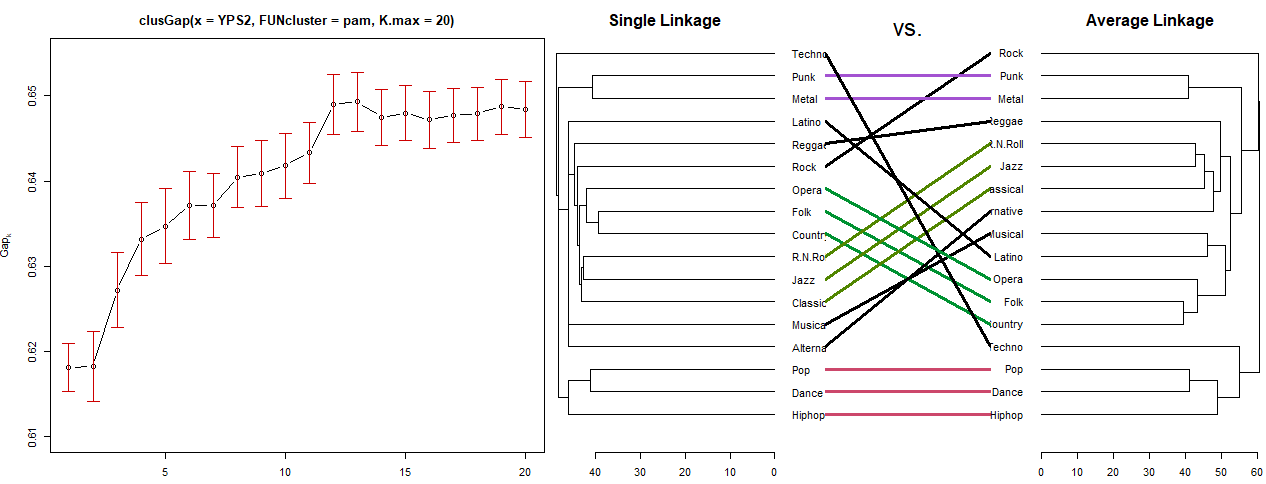}
\caption{Analysis of Youth Music, Movie, and Activity Preferences}
\label{fg:ClusterPlot}
\end{figure}

R packages for classification and grouping are listed in Table \ref{tb:RSegmentation}.  A range of both classical and mixture model/latent class based procedures are included.  Packages that have features for visualization and cluster validation are included. Procedures designed for large datasets and datasets with mixed measurement types are also included. 
\begin{table}[h]
\centering
\caption{Segmentation and Grouping Packages in R}
\label{tb:RSegmentation}
\begin{tabular}{p{1.6cm} p{3.4cm} p{10.1cm}} \hline
\textbf{CRAN}&\textbf{Reference}&\textbf{Description}\\  \hline
base&None&hclust() implements hierarchical clustering with a range of linkage methods.\\
base&None&kmeans() implements basic k-means clustering procedure.\\
CluMix&\citet{REF1425}&Cluster analysis and visualization for mixed (continuous and categorical) data.\\
cluster&\citet{REF1366}&General clustering package. Contains clara() for large scale k-means clustering and pam() for partitioning on medoids.\\
dendextend&\citet{REF1423}&A package for visualizing and comparing dendograms from hierarchical clustering.\\ 
fastclust&\citet{REF1436}&Fast, scalable hierarchical clustering algorithms for large datasets.\\
fclust&\citet{REF1424}&Implements fuzzy clustering algorithms.\\
flexclust&\citet{REF1367}&K-centroids clustering, with a choice of distance metrics and various advanced methods such as neural clustering.\\
fpc&\citet{REF1362}&Contains clusterwise regression, a range of fixed point clustering methods, and clustering validation routines.\\
MCLUST&\citet{REF1365}&Gaussian model based clustering models and algorithms.\\
poLCA&\citet{REF1363}&Polytomous variable latent class analysis, including latent class (clusterwise) regression.\\ 
skmeans&\citet{REF697}&Contains an interface to the CLUTO vmeans function which provides a range of criteria for partitioning clustering from distance matrices. Particularly useful for text/document clustering.\\ \hline
\end{tabular}
\end{table}
An example is given in Figure \ref{fg:ClusterPlot}, which illustrates the results from cluster analysis performed with the previously utilized youth preferences dataset.  The first subplot utilizes the ``cluster'' package to give the gap statistic for $k=1..20$ k-medoids clustering solutions for partitioning the survey participants.  Using the algorithm described by \citet{REF686}, a 12 cluster solution is suggested. The second subplot uses the ``dendextend'' package and  the hclust() function to compare single and average linkage clustering solutions for the different music genres.  There are some differences between the solutions, but also some commonalities.  As with the previous visualization solution, there is a degree of face validity, particularly for the average linkage solutions.  The sets of genres of \{pop, dance, hip-hop\}, \{punk, metal, rock\}, and \{folk, country\} are clustered together in groups towards the bottom of the tree, indicating close correspondence. 

\subsection{Conclusions}
Segmentation, as a method, has been utilized by both marketing academics and practitioners. Segmentation methods remain important for marketing practitioners.  In fact, in a recent survey designed to determine the influence of major marketing science innovations on marketing practice \citep{REF1571}, segmentation was the highest rated innovation by both academics and practitioners.  However, despite this overall impact, in a critical review of market segmentation in the Harvard Business Review, \citet{REF1357} described a survey of 200 senior executives, where 59\% reported carrying out a major segmentation exercise within the last two years, but only 14\% reported that any real business value resulted from the exercise.  The authors note that psychographic segmentation has become widely established over the last 50 years and has been instrumental in some foundational advertising campaigns, including the Pepsi ``Generation'' campaign, which melded together groups of consumers who identified with youth culture.  However,  the authors note that psychographic segmentation still has limited usefulness when it comes to predicting specific brand behavior.  This ties in with some of the issues previously raised by \citet{REF1302}  and the trade-offs between general behavioral and brand specific attributes. Furthermore, \citet{REF1361} note that despite technical advances in segmentation in the 1980s and 1990s, segmenting towards needs or feelings rarely finds actionable segments of customers who can be targeted in the real business world.  To counter this and other issues, the authors recommend a multidisciplinary team to brainstorm segmentation criteria and to paying attention to a broad range of factors including the purchase/user environment, a customer's desired experience, and product beliefs/associations, in order to be able derive segments related to purchase and usage behavior.  

Several recent advances hold promise in the segmentation arena. The first is the use of neuroscience, which provides a set of segmentation bases based on unconscious cognitive responses to stimuli \citep{REF1360}, can provide brand-specific psychological responses from consumers, and when combined with traditional segmentation bases, can provide great insight into brand behavior.  Another advance is the rise of micro-segmentation. A New York Times article \citep{REF1586} describes how the BlueKai platform (now owned by Oracle), which contains a wide variety of behavioral and financial data from a range of sources, including browsing behavior, credit card records, and e-commerce purchases, can be used to group consumers into small micro-segments based on behavior and demographics.  Examples given include ``Hawaii-vacation-seeking price-sensitive Democrat'' and ``baseball-loving safety-net senior oenophile''.  These segments contain quite specific product needs and can be algorithmically targeted by flexible e-commerce engines. 

\section{Class Prediction}
\subsection{Discriminant Analysis and Related Techniques}
Marketing prediction encompasses a plethora of models for a range of responses, including purchase behavior, review ratings, customer loyalty, customer lifetime value, sales, profit, and brand visibility. To keep length manageable we will concentrate on methods of prediction where predictions are made for either a specific class label and for which the scope of the prediction is at a high level of ``granularity'', i.e., predictions for an individual consumer or product.  For example, for a customer renewal prediction application, let $y_i\in\{0,1\}$, where 0 indicates that a customer ``churned'' or did not renew a contract and $y_i=1$ indicates the converse.  Class prediction techniques can be applied to multi-class classification problems where there are more than two categories and also instances where there are more than two labels to be predicted, the so-called multi-label classification problem \citep{REF627}.  Though class prediction requires discrete data, class prediction methods can be applied to discretized continuous data; for example, \citet{REF1418} operationalize change in service utilization as a dichotomous increase/decrease variable.

In most instances, models will be built and tested using data for which $y_i$ is known and then applied on data for which $y_i$ is not known. From a data mining terminology standpoint, this is known as ``supervised learning'', as opposed to the unsupervised learning techniques of dimensionality reduction and cluster analysis, where there is no known dependent variable that can be used to guide the output.  Typically, models will be tested with some type of holdout or cross validation scheme \citep{REF1372}.  Here, a proportion of the dataset is denoted as the training data and is used to predict $y_i$  on the remaining holdout or test data.  This can be repeated iteratively by repeatedly sampling test data without replacement until every observation is included in a holdout dataset.  If \textit{p} items are held out at each iteration then the procedure is known as ``hold-p-out cross validation''.  If a proportion of $1/n$ of the data is held out at each iteration then the procedure is known as ``n-fold cross validation''.  The predictions for the test data $\hat{y}_i$ will then be compared against the actual data $y_i$, either using a cross classification table (for class label predication), and/or some measure of error, for example the mean squared error (MSE), where $MSE=\frac{1}{n}\sum_{i=1}^n\left(\hat{y}_i-y_i\right)^2$.

Most early class prediction papers in marketing utilized discriminant analysis \citep{REF1379}.  In discriminant analysis, each item has a set of features ${\bf{x}}_i$ and a class label $y_i$.  The data are split into groups using the class label $y_i$. In the binary case, a discriminant function line is drawn between the groups to maximize the variance between groups relative to the variance within groups. This ``discriminant function'' is defined as a weighted combination of the features of ${\bf{x}}_i$, so if ${\bf{w}}'{\bf{x}}_i>c$ the item \textit{i} is classified into one group and if ${\bf{w}}'{\bf{x}}_i\leq c$, the item is classified into the other group.  An initial application of discriminant analysis in marketing was to predict brand switching behavior \citep{REF1380} (switch/not switch) using predictors of quantity purchased, income, and family size.  \citet{REF1381} describe applications for multiple discriminant analysis, where there may be more than two classes, e.g., one wishes to predict consumer choice between three brands {A,B,C}. They note that prediction performance on the sample may lead to overall bias and recommend either using holdout validation or adjusting the prediction success using the success predicted by chance.  Further examples of the use of discriminant analysis include the prediction of consumer innovators \citep{REF1382}, new product purchasers \citep{REF1383}, choice of retail facilities \citep{REF1384}, business to business source loyalty \citep{REF1386} and private label brands \citep{REF1385}. Issues surrounding the use of discriminant analysis are very similar to those found in market segmentation, including the selection of analysis variables and interpretation of results.  \citet{REF1387} notes that the probability of group membership calculated by discriminant analysis is a product of its likelihood ratio and its prior odds, so for a two label problem where $y_i\in\{0,1\}$, the ratio of probabilities for group membership is given in \eqref{eq:Discriminant}.
\begin{equation}
\label{eq:Discriminant}
\frac{p\left(y_i=1|{\bf{x}}_i\right)}{p\left(y_i=0|{\bf{x}}_i\right)}=\frac{p\left({\bf{x}}_i|y_i=1\right)}{p\left({\bf{x}}_i|y_i=0\right)}\times\frac{p\left(y_i=1\right)}{p\left(y_i=0\right)}
\end{equation}
Here, the first term is the posterior ratio of group membership probabilities, the second term is the likelihood ratio, and the third term is the ratio of prior probabilities, which are operationalized as the proportion of items in each class. The log of likelihood ratio is the discriminant function ${\bf{w}}'{\bf{x}}_i$, which links discriminant analysis to logistic regression. \citet{REF1387} further notes that the ratio of membership probabilities can be further altered by the economic cost of misclassifying items.  For uneven classes, the ratio of prior probabilities can dominate the likelihood ratio and thus lead to all the items predicted to be in the larger class. A possible solution is to assign the items closest to the decision boundary to the smaller class, thus damping the effect of the uneven class memberships. The weight coefficients of the discriminant function can be interpreted in a similar way to regression.  If the variables are not standardized then the coefficients give the absolute effect of one unit of each of the elements of ${\bf{x}}_i$ on the discriminant function.

\subsection{Trends in Data Mining and Statistical Learning}
Over the subsequent decades, a whole host of algorithms have been developed for class prediction.  Many of these give better performance on certain problems than linear discriminant analysis.  However, at their core, class prediction algorithms have two main purposes.  The first is to ``predict'' the class labels and the second is to explain how the item (customer etc.) features inform the prediction result.  Consider a brand loyalty example.  Marketers may wish to pay extra attention to those customers who are predicted to churn.  In the context of planning a marketing campaign, the actual prediction probabilities give more information than the class labels and may help with resource allocation.  In addition, parameter estimates can be used to help relate  ${\bf{x}}_i$ to the predicted $y_i$ and give a broader context to the prediction decision. 

The method of support vector machines (SVMs) \citep{REF1388} has has found wide applicability in marketing prediction problems and can give performance increases over the logit/probit models that are often used in marketing prediction applications \citep{REF1389}.  The intuition for SVMs is very similar to discriminant analysis.  Here, $y_i\in\{-1,1\}$ and a boundary line (or hyperplane) between classes is defined as ${\bf{w}}'{\bf{x}}_i-c=0$.  The value of ${\bf{w}}$ is then optimized using \eqref{eq:SVM}. The inner maximization equation tries to keep separation between classes, by defining a margin of size 1 and penalizing items that fall within the margin.  The value of $\lambda$ provides a trade off as to how strongly the margin is enforced.
\begin{equation}
\label{eq:SVM}
\min \left[\frac{1}{n}\sum_{i=1}^{n}max\left(1-y_i\left({\bf{w}}'{\bf{x}}_i-c\right),0\right)\right]+\lambda\left\|{\bf{w}}\right\|^2
\end{equation}
Both discriminant analysis\footnote{discriminant analysis, also has a quadratic variant, which occurs when the group variances are assumed to be unequal} and SVMs utilize distances and the dot products in the distances $ \langle{\bf{x}},{\bf{x}}\rangle$ can be transformed by a ``kernel'' to give a nonlinear boundary between decision groups.  For discriminant analysis and SVM, this ``kernel trick'' \citep{REF444}, gives rise to non-linear decision boundaries and by selecting the kernel type (Gaussian, polynomial, linear, etc.) and tuning parameters, strong performance can be gained from a wide range of datasets. 

A wide range of other methods have been applied to class prediction problems in marketing.  These include probabilistic tree methods, such as CAR \citep{REF1394} and random forests \citep{REF1391}, neural networks \citep{REF1392}, nearest neighbor methods \citep{REF1393}, stochastic gradient boosting \citep{REF1402}, and na\"ive Bayes \citep{REF1395}.  For parsimony, full descriptions are left out in this review, but there are many excellent resources available, for example, \citet{REF257}, who discuss classification techniques from a statistical learning standpoint and \citet{REF1396}, who follow a purer machine learning/pattern recognition approach.

There are several major trends in class prediction.  The first is the rise in the use of so called ``ensemble'' methods, where results from multiple algorithms are combined to give better results. This approach gets around the problem that for class prediction applications, relative algorithm performance can be dataset specific.  In fact, several of the aforementioned techniques, random forests and stochastic gradient boosting, are predicated on combining multiple weak classifiers into a strong classifier. Another development has been for on-line prediction competitions on platforms such as \url{Kaggle.com}, where datasets from industry/science are uploaded for a specific prediction task, with competitors building models on training data and then having their solutions validated on a hidden test dataset.  In this realm of practical prediction analysis, the best solutions often are ``ensemble'' solutions. Common methods for creating ensemble solutions include the statistical pooling techniques of  ``boosting'' and ``bagging'' \citep{REF1399,REF1400,REF1401}. In addition, neural networks can be used to find the best set of weights for combining solutions \citep{REF1407}. 

A second trend is the use of feature reduction and extraction techniques.  Modern datasets can be extremely complex with hundreds of thousands of features or dimensions, many of which are mainly noise and don't improve prediction accuracy.  Feature selection methods can improve the signal to noise ratio and reduce the size of the dataset, which improves scalability.  The previously described dimensionality reduction technique of PCA, along with related methods, can be used as a precursor to or combined with class prediction techniques such as SVM \citep{REF1403}. An alternate approach is to select the most useful features.  This can be done using a wide range of criteria including high correlations with the class labels, low correlations with other predictor variables, predictor performance for single variable classifiers, and information theoretic measures \citep{REF1404}.  There are many bespoke feature selection algorithms for specific types of data.  For example, in an interesting application, \citet{REF1416} take binary brand purchase data utilize the ``pick-any'' joint space mapping technique previously referenced \citep{REF1417}, and use the output dimensions from this technique to predict brand purchases.

\subsection{Churn Prediction}

Recent marketing academic interest in class prediction techniques has been spurred by the availability of data in CRM systems \citep{REF491}. \citet{REF259} describe the results of a tournament, where both academics and practitioners used a training dataset to build models for churn prediction, which were then evaluated on a test dataset.  Several success metrics were reported, which all rely on items being ordered in order of the probability of churn, i.e. $p\left(y_i=1\right)$. The decile lift gives the ratio of the proportion of churners in the top 10\% of predicted churners to the proportion in the full dataset. The GINI coefficient, adapted from the GINI measure of economic inequality, gives a measure of how unequally distributed the actual churners are on the ordered list.  The authors note that logit based models performed well on the dataset and outperformed discriminant and tree based techniques.  The data mining aspects of churn management are linked to marketing work in CLV or customer lifetime value \citep{REF1354}, to produce a profitability metric for a churn management campaign, where potential customers are offered an incentive to remain, which is given in \eqref{eq:CLVProfit}.
\begin{equation}
\label{eq:CLVProfit}
Profit=N\alpha\left[\left(\gamma CLV+c_{IN}\left(1-\gamma\right)\right)\beta_{0}\lambda-c_{CO}-c_{IN}\right]-c_{FX}
\end{equation}
Here, $\beta_0$ is the proportion of churners, $\lambda$ is the ``lift'', i.e., the proportion of churners for the targeted customers divided by the proportion of churners across all customers, $\gamma$ is the success of the incentive, i.e., the proportion of targeted consumers who remain loyal, $c_{CO}$ is the cost of contacting consumers, $c_{IN}$ is the cost of the incentive, and $c_{FX}$ are fixed promotion costs.  An extension is given by \citet{REF1408}, who model profit using a parameterized beta distribution and from this create an EPMC (expected profit maximization criterion) as an evaluation metric for churn prediction, which can be used for feature selection \citep{REF1420}.

Many of the described churn prediction algorithms can be applied to other scenarios, for example customer targeting prediction \citep{REF1415} or yes/no recommendation prediction.   With churn data, there can be a strong class imbalance problem, with only a few churners and many non-churners.  Predictive accuracy can be improved either by under sampling non-churners \citep{REF1405}  or by oversampling churners \citep{REF1592,REF1593}.  As described previously, the usefulness of class prediction algorithms is predicated on both prediction accuracy and the ability to interpret model parameters. \citet{REF1406} describe a framework based on an ensemble of additive logit models and build in a set of feature importance scores, to help interpret/select features. They implement graphs built using splines that show the probability function of churning and associated confidence intervals alongside a histogram of churn class distribution.

Corporate data utilized for churn prediction models have been annotated with derived sentiment data from emails \citep{REF1411} and from information available from company websites \citet{REF1409}. Given increasingly strict data protection laws, particularly in Europe, it is sometimes necessary to delete past customer data, rendering it unamenable to analysis.  To get around this problem, \citet{REF1410} describe a method that only needs model parameters to estimate prediction models and uses Kalman filters to update the parameters as new data come in.  Another issue is that training and implementation datasets may have differing distributions due to rapid changes in the business environment due to biased sampling. To deal with the first situation, \citet{REF1419} define and optimize a time window for which customer events are included in the prediction model. \citet{REF1413} start with the assumption that training and test datasets have different distributions and employ a transfer learning \citep{REF1412} algorithm, which uses a neural network to match features between training and test datasets and combines this process with an ensemble classifier.  To be used in practice for large scale consumer datasets, churn prediction must be scalable. \citet{REF1414} describe the deployment of a large scale churn prediction system for a Chinese mobile operator with millions of active customers and multiple large databases, containing 2.3 terabytes of information.  To achieve scalability, a tiered architecture was used with a data layer containing the databases, a ``big data'' layer consisting of data integration, modeling, and mining components, and an applications layer containing the business process functions, including churn prediction. The system is analyzed with respect to the volume, velocity, and variety of the data.

\subsection{Software}
\begin{figure}[h]
\centering
\includegraphics[width=16cm]{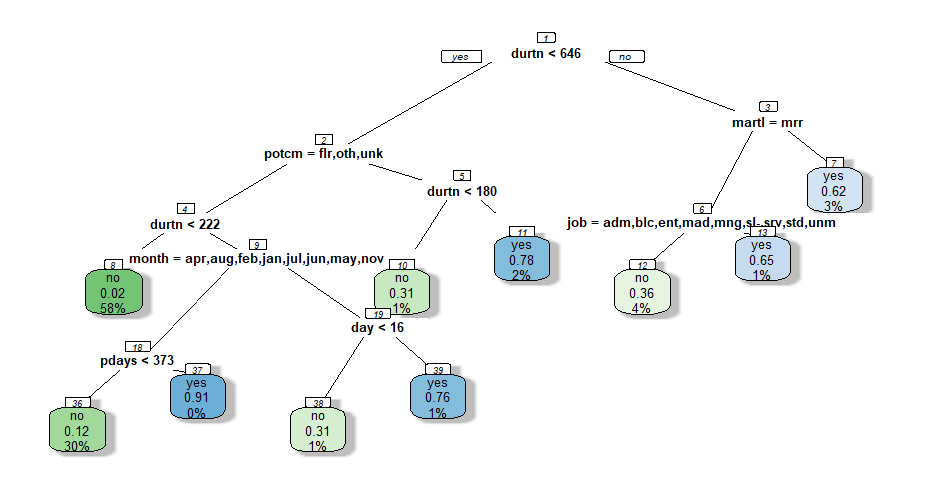}
\caption{Classification Tree for Bank Marketing Promotion (yes/no)}
\label{fg:RPartTree}
\end{figure}

R packages for class prediction are listed in Table \ref{tb:RPrediction}.  Several general classification packages are listed, including caret and RWeka, which both provide fully fledged environments for designing, building, and testing class prediction implementations. Basic discriminant analysis and SVM techniques are listed, along with implementations of tree methods and ensemble methods. The basic two class logit and probit models are included in the base R libraries.  The mlogit and mprobit libraries are given for the multinomial logit and probit techniques.
\begin{table}[h]
\centering
\caption{Segmentation and Grouping Packages in R}
\label{tb:RPrediction}
\begin{tabular}{p{1.6cm} p{3.4cm} p{10.1cm}} \hline
\textbf{CRAN}&\textbf{Reference}&\textbf{Description}\\  \hline
asa&\citet{REF1444}&Stochastic gradient boosting classification.\\
base&None&glm() implements logistic regression with family=binomial, binary logit with family=binomial(link=``logit''), and binary probit with family=binomial(link=``probit'').\\
Boruta&\citet{REF1454}&Feature selection using random forests to remove unimportant features.\\
caret&\citet{REF1439}&Access to class prediction algorithms, feature selection methods, and cross-validation procedures to evaluate performance. caretEnsemble allows an ensemble of caret classifiers.\\
class&\citet{REF1437}&Contains a range of methods including SOMs, learning vector quantization and k-nearest neighbours.\\
e1071&\citet{REF1440}&Contains functions to implement SVMs and includes a choice of kernels.\\
FSelector&\citet{REF1455}&Feature selection, including correlation, entropy, and chi-squared based methods.\\
MASS&\citet{REF1437}&Contains lda() routine for linear discriminant analysis and qda() routine for quadratic discriminant analysis.\\
MCLUST&\citet{REF1365}&Guassian model based clustering models and algorithms.\\
mlogit&\citet{REF1436}&Implements a multinomial logit model.\\ 
mprobit&\citet{REF1436}&Implements a multinomial probit model.\\ 
neuralnet&\citet{REF1445}&Train and implement backpropogation neural networks.\\
randomForest&\citet{REF1443}&Creates random forest ensembles of weak classifiers.\\ 
ROSE&\citet{REF1597}&Implements sampling methods for imbalanced class prediction.\\ 
rpart&\citet{REF1441}&Functions to build classification (and regression) trees.\\ 
RWeka&\citet{REF1144}&An R interface to Weka, a comprehensive data cleaning, feature selection, and prediction package.\\ \hline
\end{tabular}
\end{table}
An example is given in Figure \ref{fg:RPartTree}, which shows a classification tree created with ``rpart'' for a bank marketing dataset\footnote{Dataset available at \url{https://archive.ics.uci.edu/ml/datasets/bank+marketing}}. The dependent variable is a response to a promotion (yes,no).  The independent variables used to build the tree include duration of time the customer has held a bank account, job, martial status, month, and day of the month.  At each tree branch, yes values are branched to the left and no values are branched to the right. The values displayed at each terminal node are the predicted response, the probability of a ``yes'' response, and the proportion of training instances captured at the node.  For example, node 7 gives people whose last contact with the bank was at least 646 seconds long and who are married.  These customers constitute 3\% of the customer base and have a 62\% chance of a ``yes'' response to the promotion.

\subsection{Conclusions}
There has been an explosion of interest in class prediction algorithms as of late.  Competition websites such as \url{Kaggle.com} and \url{TopCoder.com}, along with conferences such as KDD, regularly host class prediction competitions in which academics, industry practitioners, and hobbyists compete against each other in order to gain the winning prediction on a test dataset with models built from supplied training data.  The rapid growth of data science as a field and its relative immaturity has contributed to this trend, with people trying to hone their skills through competition. Class prediction methods have been utilized in marketing from the 1960s onwards and have been applied to a range of problems, particularly in the areas of customer churn and predication. With recent academic interest in big data and analytics, there has been something of a revival in this area in marketing, spurred on by \citet{REF259} and other recent work.

A few salient points can be elicited from this review.  First, there is no one ``silver bullet'' when it comes to choosing a prediction technique.  Algorithm performance varies across dataset type and characteristics.  When implementing a new prediction application, it is advisable to not only examine past work on similar datasets, but also to cast a wide net and look at a range of different algorithms. Second, ensemble methods, which combine multiple classifiers into a single classifier tend to perform very well on a range of datasets and if computational power allows, make good starting points for algorithm implementation. Third, for large complex datasets with noisy data, some sort of feature selection may be required to keep the signal to noise ratio high and keep computational costs reasonable.  Fourth, judging algorithms solely on prediction performance on a holdout dataset is taking a rather narrow view of performance.  A very tiny incremental increase in prediction percentage may only improve profit by a tiny amount and does not take account of other factors, such as the algorithm runtime, the cost of implementing the algorithm, algorithm robustness, and the interpretability of the results.  For example, ``The Netflix Prize'', though not strictly a class prediction exercise, was one of the first large-scale online data competitions. A prize of \$1,000,000 was awarded to the winning entry. Competition entrants needed to predict movie review scores for users based on their review scores for other movies.  The final solution was an amalgamation of the many different techniques tried and shared among competition entrants.  However, despite the fanfare, the winning solution was was never actually implemented.  An entry in the Netflix tech blog \citep{REF1421} notes that ``the additional accuracy gains that we measured did not seem to justify the engineering effort needed to bring them into a production environment'' and that with a move towards online streaming content the style and format required for the recommendations had changed, which illustrates that empirical prediction performance is only one aspect of the overall system needs of a marketing analytics application.

\section{Overall Discussion}
We have presented a review of marketing analytics that primarily covers the topics of visualization, segmentation and grouping, and class prediction. These topics were chosen as not only are they core to marketing strategy and have a long history in academic marketing, but are also of interest to researchers in expert systems, data mining, statistics, and operations research. There is a commonality throughout all three areas.  In the 1960s and early 1970s there were a number of papers that took methodology from statistics and psychology and applied it to managerial marketing problems of positioning, segmentation, and response prediction.  A core group of researchers in applied psychometrics and measurement, including J. Douglas Carroll, Ronald Frank, Paul Green, and Donald Morrison developed methods in all three of the fundamental areas of visualization, segmentation, and class prediction.  This was part of an overall growth in interest in quantitative marketing in the 1960s, spearheaded by researchers such as Frank Bass and Andrew Ehrenberg who produced pioneering work in the areas of product diffusion \citep{REF769} and consumer choice \citep{REF20,REF1422}. The rapid growth of quantitative marketing in this era was spurred on by the availability of computational tools and data, an initiative by the Ford Foundation to equip business faculty with skills in mathematics and quantitative analysis, and the founding of the Marketing Science Institute to support the application of scientific techniques to marketing \citep[pp. 10--15]{REF1573}.

There has been a steady stream of methodological work in the intervening period, but the last ten years have seen an explosion of interest in marketing analytics for several reasons.  The first of these is the availability of data.  Many of the papers cited in this review give extensions of basic methods, but are designed to deal with large, complex modern datasets, such as on-line reviews (including text), web-logs, CRM data, and brand scanner data.  Much of this work is cross disciplinary, with researchers in different fields working with one another and citing one others' work.  This has partly been brought about by increased computational sophistication in statistics and increased interest in statistical methodology in expert systems and data mining.

Marketing models and methods do not exist in a vacuum.  Most are implemented within computational software or systems.  In fact, many advances in marketing analytics methodology and many innovative applications are published in the expert and intelligent systems literature, including in this journal.  Our hope is by publishing this review in an expert systems journal and positioning it at the intersection of multiple fields, we have achieved several things.  First, we have provided a historical context and managerial marketing insight to researchers in the expert systems field.  Second, as much applied marketing segmentation research is now carried out in the expert systems field, we will have made researchers in the marketing domain aware of this work.  Third, by including more theoretical work from statistics and operations research, some of this work can filter through to applied researchers.

Given the challenges and sophistication of modern data analysis problems, it is our view that this interdisciplinary approach will continue and strengthen over the coming decades, as solving practical problems will require a range of computational, statistical, and business skills.  While marketing analytics implementations can achieve strong return on investment for businesses, success is correlated with a strong analytics infrastructure and culture, elements that need buy-in from multiple areas of a company, including marketing, IT, and senior decision makers \citep{REF1570}.

Several authors have noted a disconnect between academics and practitioners in marketing, due to increased specialization and siloing of research \citep{REF1572}, academic incentives that reward publications in academic journals rather than broader commercial impact \citep{REF1574}, and a hesitancy on the part of academics to engage with practitioners \citep{REF1575}.  However, outside of marketing, there is less worry about this disconnect, possibly as large numbers of Ph.D. graduates in areas such as statistics and expert systems go into industry, which helps narrow any academic-practitioner disconnect. We predict that there will be increasingly close interactions between academics and practitioners in marketing analytics. There are several reasons for this. First, increasing numbers of Ph.D. graduates are going into analytics/big data jobs in industry, where technical skills are at a premium, thus leading to an overlap in professional networks between academia and industry. Second, there has been a concerted effort from academics to engage with businesses and solve industry problems.  Examples of this include the Wharton Customer Analytics Initiative, which works with industry partners to provide datasets and associated research opportunities for researchers, and well regarded practice prizes from the INFORMS and INFORMS marketing science communities, which are designed to reward research that has strong real world outcomes and impact \citep{REF1569}. Third, business schools are increasingly emphasizing analytics at all levels of the curriculum and a range of pedagogical material has been developed to meet this need.  For marketing analytics, in addition to classic books on marketing models \citep{REF1576} and marketing engineering \citep{REF295}, there are recent books on implementing marketing analytics using R \citep{REF1577}, building spreadsheet models for marketing analytics \citep{REF1578}, and on marketing strategy aspects of data analytics \citep{REF1579}. Fourth, as much of the technology associated with big data and analytics is new, there is a scramble to learn new techniques and methods, both from practitioners and academics.  This has lead to a range of meet-up groups that are targeted to people wishing to learn new technologies, which attract both practitioners and academics. This phenomenon, along with the growth of data science competitions, which frequently feature marketing data, and the use of open-source software such as R and Python, has lead to a vibrant marketing analytics community, which encompasses both practitioners and academics. This bodes well for the future.

\section{Acknowledgements}
This research did not receive any specific grant from funding agencies in the public, commercial, or
not-for-profit sectors.

{\small \singlespace \bibliography{MarketingAnalytics4}}
\end{document}